\documentclass{article}

\PassOptionsToPackage{numbers}{natbib}

\usepackage[preprint]{neurips_2026}

\usepackage[utf8]{inputenc} 
\usepackage[T1]{fontenc}    
\usepackage{hyperref}       
\usepackage{url}            
\usepackage{booktabs}       
\usepackage{amsfonts}       
\usepackage{nicefrac}       
\usepackage{microtype}      
\usepackage{xcolor}         

\usepackage{authblk}
\usepackage{amsmath}
\usepackage{mathtools}
\usepackage{algorithm}
\usepackage{algorithmicx}
\usepackage{algpseudocode}
\usepackage{tabularx}
\usepackage{multirow}
\usepackage{makecell}
\usepackage{array}
\usepackage{subcaption}
\usepackage{textcomp}
\usepackage{stfloats}
\usepackage{verbatim}
\usepackage{graphicx}
\usepackage{color}
\usepackage{glossaries}
\usepackage{enumitem}
\usepackage{wrapfig}

\glsdisablehyper
\setacronymstyle{long-short}
\newacronym{ai}{AI}{Artificial Intelligence}
\newacronym{ml}{ML}{Machine Learning}
\newacronym{dl}{DL}{Deep Learning}
\newacronym{ann}{ANN}{Artificial Neural Network}
\newacronym{relu}{ReLU}{Rectified Linear Unit}
\newacronym{cnn}{CNN}{Convolutional Neural Network}
\newacronym{snn}{SNN}{Spiking Neural Network}
\newacronym{csnn}{CSNN}{Convolutional Spiking Neural Network}
\newacronym{if}{IF}{Integrate-and-Fire}
\newacronym{lif}{LIF}{Leaky Integrate-and-Fire}
\newacronym{ssif}{SSIF}{Single-Spike Integrate-and-Fire}
\newacronym{relpsp}{ReL-PSP}{Rectified Linear Postsynaptic Potential}
\newacronym{ttfs}{TTFS}{Time-to-First-Spike}
\newacronym{wta}{WTA}{Winner-Takes-All}
\newacronym{pcn}{PCN}{Paired Competing Neurons}
\newacronym{ncg}{NCG}{Neuronal Competition Groups}
\newacronym{bp}{BP}{Backpropagation}
\newacronym{gd}{GD}{Gradient Descent}
\newacronym{bptt}{BPTT}{Backpropagation Through Time}
\newacronym{stdp}{STDP}{Spike Timing-Dependent Plasticity}
\newacronym{rstdp}{R-STDP}{Reward-Modulated STDP}
\newacronym{sstdp}{SSTDP}{Supervised STDP}
\newacronym{s2stdp}{S2-STDP}{Stabilized Supervised STDP}
\newacronym{fa}{FA}{Feedback Alignment}
\newacronym{sfa}{sFA}{Sign-concordant Feedback Alignment}
\newacronym{dfa}{DFA}{Direct Feedback Alignment}
\newacronym{ss}{SS}{Sign-Symmetry}
\newacronym{srm}{SRM}{Spike Response Model}
\newacronym{psp}{PSP}{Post-Synaptic Potential}
\newacronym{dft}{DFT}{Dynamic Firing Threshold}
\newacronym{kp}{KP}{Kolen-Pollack}
\newacronym{wm}{WM}{Weight Mirror}
\newacronym{fbp}{\textsc{fBP}}{Frozen Backpropagation}
\newacronym{fbpall}{\textsc{fBP}}{Frozen Backpropagation}
\newacronym{fbpproba}{\textsc{proba}}{Frozen Backpropagation with Proba}
\newacronym{fbprandom}{\textsc{random}}{Frozen Backpropagation with Random}
\newacronym{fbpquantile}{\textsc{quantile}}{Frozen Backpropagation with Quantile}

\newcolumntype{C}{>{\centering\arraybackslash}X}

\newcommand{\glsdef}[1]{\glsentrylong{#1} (\glsentryshort{#1})}

\title{Frozen Backpropagation: Relaxing Weight Symmetry in Deep Spiking Neural Networks}

\author[1]{Gaspard Goupy}
\author[2]{Pierre Tirilly}
\author[2,*]{Ioan Marius Bilasco}
\affil[1]{Nokia Bell Labs, Espoo, Finland}
\affil[2]{Univ. Lille, CNRS, Centrale Lille, UMR 9189 CRIStAL, F-59000 Lille, France}
\affil[*]{Corresponding author: marius.bilasco@univ-lille.fr}

\begin{document}

\maketitle

\begin{abstract}
    Direct training of Spiking Neural Networks (SNNs) on neuromorphic hardware can greatly reduce energy costs compared to GPU-based training.
    However, implementing Backpropagation (BP) on such hardware is challenging because forward and backward passes are typically performed by separate networks with distinct weights.
    To compute correct gradients, forward and feedback weights must remain symmetric during training, necessitating weight transport between the two networks.
    This symmetry requirement imposes hardware overhead and increases energy costs.
    To address this issue, we introduce Frozen Backpropagation (\textsc{fBP}), a BP-based training algorithm relaxing weight symmetry in settings with separate networks.
    \textsc{fBP} updates forward weights by computing gradients with periodically frozen feedback weights, reducing weight transports during training and minimizing synchronization overhead.
    To further improve transport efficiency, we propose three partial weight transport schemes of varying computational complexity, where only a subset of weights is transported at a time.
    We evaluate our methods on image recognition tasks using both temporally and rate-coded SNNs, and compare them to existing approaches addressing the weight symmetry requirement.
    Our results show that \textsc{fBP} outperforms these methods and achieves accuracy comparable to BP while significantly lowering transport costs.
    With partial weight transport, \textsc{fBP} can further lower those costs by up to \(10{,}000\times\) at the expense of moderate accuracy loss. 
    This work provides insights for guiding the design of neuromorphic hardware incorporating BP-based on-chip learning.
\end{abstract}

\section{Introduction} \label{chap:fbp:sec:intro}
\glspl{snn} are a promising alternative to second-generation \glspl{ann} for energy-efficient computing~\citep{yamazakiSpikingNeuralNetworks2022}.
Coupled with neuromorphic hardware~\citep{schumanSurveyNeuromorphicComputing2017}, which supports highly parallel processing and in-memory computing, \glspl{snn} can overcome the von Neumann bottleneck and reduce energy consumption by orders of magnitude compared to conventional CPU/GPU-based platforms~\citep{blouwBenchmarkingKeywordSpotting2019,ostrauBenchmarkingDeepSpiking2020,fedorovaAdvancingNeuralNetworks2024}.
To unlock their energy-efficient potential and facilitate hardware deployment, neuromorphic chips should incorporate scalable on-chip learning~\citep{stanojevicHighPerformanceDeepSpiking2024}.

\begin{figure}[t]
    \centering
    \includegraphics[width=0.95\linewidth]{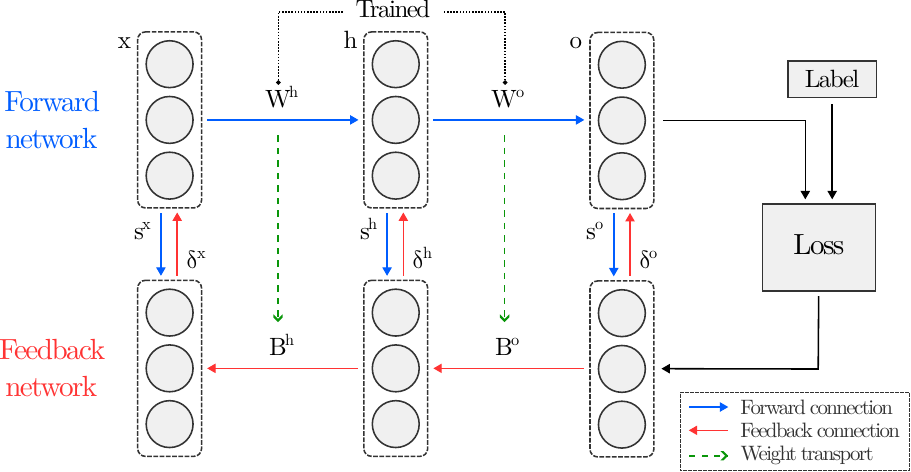}
    \caption{\gls{bp}-based training in a dual-network configuration, consisting of a forward and a feedback network with distinct, unidirectional synapses. The forward network uses weights \(W\) to compute neuron activations \(s\). The feedback network uses weights \(B\) to compute neuron errors \(\delta\). Neuron errors act as feedback signals triggering weight updates only in the forward network. For correct gradient computation, weight transport is needed to maintain \(B\) symmetric with \(W\) during training.}
    \label{chap:fbp:fig:dual-network}
\end{figure}

State-of-the-art training methods for \glspl{snn} are based on \gls{bp}~\citep{eshraghianTrainingSpikingNeural2023}, with \gls{bptt}~\citep{wuSpatioTemporalBackpropagationTraining2018,liSEENNTemporalSpiking2023} for rate-coded networks, and event-driven \gls{bp}~\citep{weiTemporalCodedSpikingNeural2023,stanojevicHighPerformanceDeepSpiking2024} for temporally-coded networks.
However, \gls{bp}-based algorithms remain challenging to implement on neuromorphic hardware~\citep{neftciEventDrivenRandomBackPropagation2017,zenkeBrainInspiredLearningNeuromorphic2021,rennerBackpropagationAlgorithmImplemented2024}.
One key challenge is that neuromorphic hardware typically uses unidirectional synapses (like biological systems), requiring separate networks with distinct weights for the forward and backward passes~\citep{rennerBackpropagationAlgorithmImplemented2024}.
We refer to this setup as dual-network configuration, illustrated in Figure~\ref{chap:fbp:fig:dual-network}.
In this setup, only the forward weights are updated, based on neuron errors computed with the feedback weights.
To ensure correct credit assignment, the feedback weights must remain perfectly symmetric with the forward weights during training, which requires realignment after each iteration (i.e., each update of the forward weights).
This issue, known as the weight transport problem~\citep{grossbergCompetitiveLearningInteractive1987,akroutDeepLearningWeight2019,rennerBackpropagationAlgorithmImplemented2024}, or the weight symmetry requirement~\citep{liaoHowImportantWeight2016,kuninTwoRoutesScalable2020}, is both challenging and costly due to increased circuitry and energy costs associated with data movement~\citep{craftonLocalLearningRRAM2019,zenkeBrainInspiredLearningNeuromorphic2021}.

Several alternatives to \gls{bp} have been explored to eliminate or reduce the need for weight transport~\cite{kaiserSynapticPlasticityDynamics2020,zhangTuningConvolutionalSpiking2021,mirsadeghiSpikeTimeDisplacementBased2023,goupyNeuronalCompetitionGroups2024,bachoLowvarianceForwardGradients2024}.
Among them, feedback-driven methods~\citep{lillicrapRandomSynapticFeedback2016,noklandDirectFeedbackAlignment2016,xiaoBiologicallyPlausibleLearningAlgorithms2019,akroutDeepLearningWeight2019} are particularly attractive because they enable \gls{bp}-like training.
\gls{fa}~\citep{lillicrapRandomSynapticFeedback2016} eliminates weight transport by propagating errors through fixed, random feedback matrices instead of the exact transpose of the forward weights.
\gls{fa} and its variants have been widely studied for training shallow \glspl{snn}~\citep{neftciEventDrivenRandomBackPropagation2017,zhaoGLSNNMultiLayerSpiking2020,shresthaInHardwareLearningMultilayer2021,heSTSFSpikingTime2025}.
However, studies on \glspl{ann} have shown that \gls{fa} struggles to scale to deeper networks~\citep{bartunovAssessingScalabilityBiologicallyMotivated2018,moskovitzFeedbackAlignmentDeep2019,xiaoBiologicallyPlausibleLearningAlgorithms2019}, unless sign symmetry between forward and feedback weights is enforced, as in \gls{sfa}~\citep{liaoHowImportantWeight2016,moskovitzFeedbackAlignmentDeep2019}.
To date, \gls{sfa} has only been applied to \glspl{ann}.
Alternatively, \gls{ss}~\citep{xiaoBiologicallyPlausibleLearningAlgorithms2019} offers a relaxed form of weight symmetry, where the feedback weights are reduced to the sign of the forward weights.
This approach scales better to deeper \gls{ann} networks than \gls{fa}~\citep{xiaoBiologicallyPlausibleLearningAlgorithms2019}, but it still relies on sign transport to maintain sign symmetry, similar to \gls{sfa}.
Exploration of \gls{ss} in \glspl{snn} is limited; to our knowledge, it has been used in one study only, for continual learning~\citep{xiaoHebbianLearningBased2024}.
In summary, \gls{ss} and \gls{sfa} offer simple and relatively scalable alternatives to exact weight symmetry when relying on sign transport.
However, their application to deep \glspl{snn} remains largely unexplored.
In addition, the magnitude mismatch between forward and feedback weights inevitably introduces bias, distorting the scale of the gradients.
To better approximate the gradients of \gls{bp}, there is a need for methods that relax weight symmetry while minimizing this magnitude mismatch.

In this paper, we address weight symmetry in deep \glspl{snn} under dual-network configurations.
We aim to improve training efficiency by reducing the frequency of weight transport and minimizing the synchronization overhead between the forward and feedback networks.
Our main contributions can be summarized as follows:
\begin{enumerate}[leftmargin=*]
    \item We introduce \gls{fbp}, a \gls{bp}-based training algorithm that reduces the frequency of weight transport by updating forward weights using gradients computed from frozen feedback weights. At fixed intervals, the forward weights are transported to realign the feedback weights, minimizing magnitude mismatch. Importantly, \gls{fbp} is agnostic to the specific \gls{bp} algorithm (e.g., \gls{bptt}, event-driven \gls{bp}) and network architecture.
    \item To further optimize transport efficiency, we propose a partial weight transport scheme where only a subset of the weights is transported at a time. We design three selection strategies of varying computational complexities.
    \item We benchmark \gls{sfa} and \gls{ss} for the first time in deep temporally and rate-coded \glspl{snn}. We show that \gls{fbp} outperforms these methods on image recognition tasks and achieves accuracy comparable to \gls{bp}. Also, we study the trade-off between transport efficiency and accuracy: with partial weight transport, temporally-coded \glspl{snn} trained using \gls{fbp} can lower transport costs by \(1{,}000\times\) with an accuracy drop of only \(0.5\)~pp on CIFAR-10 and \(1.1\)~pp on CIFAR-100, or by up to \(10{,}000\times\) at the expense of moderate accuracy loss.
\end{enumerate}
This work provides insights to guide future efforts in designing neuromorphic hardware with efficient \gls{bp}-based on-chip learning.
The source code is publicly available at: \url{https://gitlab.univ-lille.fr/fox/fbp}.

\section{Preliminaries} \label{chap:fbp:sec:prelim}

\subsection{BP-Based Training of SNNs}

Training deep \glspl{snn} with \gls{bp} requires computing gradients with respect to neuron activations, which can be defined in different ways depending on the learning formulation: e.g., spike times or firing rates. 
Here, we formalize gradient computation based on spike times and event-driven \gls{bp}, but the same applies to other \gls{bp} algorithms; the formulation for \gls{bptt} is provided in Appendix~\ref{chap:fbp:sec:bptt-neuron-errors}.

Training a temporally-coded \gls{snn} using event-driven \gls{bp} involves computing gradients with respect to spike times:
\begin{equation}
    \frac{\partial \mathcal{L}}{\partial W^l} =
    \underbrace{
        \frac{\partial \mathcal{L}}{\partial s^L}
        \frac{\partial s^L}{\partial s^{L-1}}
        \cdots
        \frac{\partial s^{l+1}}{\partial s^{l}}
    }_{\delta^l}
    \frac{\partial s^{l}}{\partial W^{l}},
    \label{chap:fbp:eq:grad-dldw}
\end{equation}
where \(s^l\) are the spike times (i.e., neuron activations) of layer \(l\), \(W^l\) are the weights of layer \(l\), \(L\) is the output layer, and \(\mathcal{L}\) is the loss.
We refer to \(\delta^l=\frac{\partial \mathcal{L}}{\partial s^l}\) as the neuron errors, which is the gradient of the loss with respect to the spike times of neurons in layer \(l\).
Neuron errors are the central concept to capture gradient propagation in dual-network configurations.
They correspond to the non-local part of the gradient, propagated from deeper layers.

\subsection{BP-Based Training in Dual-Network Configurations} \label{chap:fbp:sec:prelim-dualnetwork}
In a dual-network configuration (Figure~\ref{chap:fbp:fig:dual-network}), forward and backward passes are computed by two networks with distinct weights: a forward network with weights \(W\), and a feedback network with weights \(B\).
Algorithm~\ref{chap:fbp:alg:bp-dualnetwork} provides a high-level overview of how training is carried out.
Note that the forward network shares neuron activations with the feedback network to compute neuron errors, while the feedback network shares neuron errors with the forward network to compute weight changes.

The weight symmetry requirement for correct gradient propagation demands weight transport from the forward to the feedback weights.
This transport is performed by copying \(W\) to \(B\) after each iteration (i.e., after each update of \(W\)), implemented through the \textsc{WeightTransport} function in step~\ref{chap:fbp:alg:step-wt}.
Several approaches have been proposed to relax this constraint; below, we formally describe two main feedback-driven methods.

\begin{algorithm}[ht]
    \caption{BP training in a dual-network configuration}
    \small
    \begin{algorithmic}[1]
        \State FN: Forward Network; BN: Feedback Network
        \For{\textbf{each} iteration \(i\)}
        \For{\(l = 1\) \textbf{to} \(L\)}
        \State (FN) Compute neuron activations \(s^l\) with \(W^l\) 
        \State (FN) Send \(s^l\) to the corresponding feedback layer
        \EndFor
        \For{\(l = L\) \textbf{to} \(1\)}
        \State (BN) Compute neuron errors \(\delta^l\) with \(B^{l+1}\) and \(\delta^{l+1}\) (or with the loss \(\mathcal{L}\) if \(l = L\))  \label{chap:fbp:alg:step-db}
        \State (BN) Send \(\delta^l\) to the corresponding forward layer
        \State (FN) Compute \(\frac{\partial s^l}{\partial W^l}\) with \(W^l\) and \(s^{l-1}\)
        \State (FN) Update \(W^l\) with \(\delta^l\) and \(\frac{\partial s^l}{\partial W^l}\) 
        \EndFor
        \State (FN) \textsc{WeightTransport}\(\left(W, B, i\right)\) \label{chap:fbp:alg:step-wt}
        \EndFor
    \end{algorithmic}
    \label{chap:fbp:alg:bp-dualnetwork}
\end{algorithm}


\paragraph{Feedback Alignment}
\gls{fa}~\citep{lillicrapRandomSynapticFeedback2016} eliminates weight symmetry by defining \(B\) as a fixed, random matrix.
To improve scalability to deeper networks, the \gls{sfa} variant~\citep{liaoHowImportantWeight2016} preserves the sign of the forward weights in the feedback path, removing weight transport but introducing sign transport instead.
In this case, \(B = \hat{B} \odot \mathrm{sign}\left(W\right)\), where \(\hat{B}\) is initialized once and fixed throughout training, \(\mathrm{sign}\left(\cdot\right)\) is the element-wise sign function, and \(\odot\) is element-wise multiplication.
This approach can be implemented by modifying the \textsc{WeightTransport} function to update the signs of \(B\) to match those of \(W\).

\paragraph{Sign-Symmetry}
\gls{ss}~\citep{xiaoBiologicallyPlausibleLearningAlgorithms2019} reduces \(B\) to the sign of \(W\): \(B = \mathrm{sign}\left(W\right)\).
Similar to \gls{sfa}, sign transport is required to maintain sign symmetry throughout training, which can be implemented by modifying the \textsc{WeightTransport} function as above.
Unlike \gls{sfa}, however, \gls{ss} eliminates stochasticity, since the feedback weights are restricted to either \(+1\) or \(-1\).


\section{Methods} \label{chap:fbp:sec:methods}

\subsection{Frozen Backpropagation} \label{chap:fbp:sec:methods-fbp}
\glsdef{fbp} is a modified \gls{bp} algorithm designed to operate in the dual-network configuration described in Section~\ref{chap:fbp:sec:prelim-dualnetwork}, with distinct forward and feedback networks.
Unlike conventional \gls{bp}, which enforces strict weight symmetry between the forward weights \(W\) and the feedback weights \(B\) through weight transport after each iteration, \gls{fbp} relaxes this requirement by decoupling the update schedules of the two networks.
Specifically, \(W\) is updated after each iteration (i.e., after each backward pass), while \(B\) is kept frozen for a fixed number of iterations \(\Phi\).
As training progresses, the symmetry between \(W\) and \(B\) is gradually broken, with neuron errors computed using a stale copy of \(W\).
After \(\Phi\) iterations, the weights are realigned by copying (i.e., transporting) \(W\) to \(B\), restoring symmetry.
This transport, illustrated in Figure~\ref{chap:fbp:fig:dual-network}, is implemented in Algorithm~\ref{chap:fbp:alg:bp-dualnetwork} by modifying the \textsc{WeightTransport} function to occur every \(\Phi\) iterations.
Therefore, \gls{fbp} reduces the frequency of weight transport during training, which may improve energy efficiency on neuromorphic hardware by reducing data movement and memory accesses.
It may also lower training latency by eliminating the synchronization overhead between forward and feedback weights after each iteration.
The hyperparameter \(\Phi\) determines the number of iterations between weight transports and thus directly translates to the transport reduction factor relative to \gls{bp}, assuming an equal number of training epochs.
For example, \(\Phi = 10\) reduces the per-epoch transport frequency by a factor of \(10\).
As such, \(\Phi\) controls the accuracy-efficiency trade-off in \gls{fbp}: larger values improve efficiency by reducing weight transport frequency, but also increase gradient bias in later iterations, which may ultimately degrade performance.
Note that \gls{fbp} is agnostic to the \gls{snn} model and the \gls{bp} algorithm, as it intervenes solely in the weight transport step.
Also, \gls{fbp} introduces no computational overhead compared to \gls{bp}.

Previous work~\citep{lillicrapRandomSynapticFeedback2016,xiaoBiologicallyPlausibleLearningAlgorithms2019} showed that exact symmetry between forward and feedback weights is not essential for scalable \gls{bp}-based learning; preserving only the sign of the weights can be sufficient.
However, such relaxed constraints may lead to suboptimal training when the weight magnitudes differ too much.
For instance, a forward weight of \(0.01\) and a feedback weight of \(0.99\) have the same sign, but the large difference causes the corresponding input neuron to be treated very differently during learning.
In the forward network, the neuron has little influence on the output spike, while in the feedback network, it is mistakenly seen as a strong contributor during error propagation.
This magnitude mismatch alters the scale of the feedback signal, leading to inaccurate gradient estimates.
The purpose of \gls{fbp} is to minimize this mismatch by periodically transporting weights from \(W\) to \(B\), ensuring that forward and feedback weights share not only the same sign but also a consistent estimate of each neuron's contribution.

\subsection{Partial Weight Transport} \label{chap:fbp:sec:methods-partial-wt}
\gls{fbp} transports weights from \(W\) to \(B\) every \(\Phi\) iterations to correct the gradient bias introduced by \(B\).
A question arises: \textit{What is the minimal number of weights that must be transported to sufficiently correct the bias?}
Transporting only a subset of the weights every \(\Phi\) iterations could further reduce transport costs and improve energy efficiency.
Based on this observation, we propose a partial weight transport scheme with three weight selection strategies of varying complexity.

\paragraph{Top-K Largest Change}
Bias arises when a weight value in \(W\) deviates too much from its counterpart in \(B\).
Moreover, larger deviations contribute to larger bias.
An effective strategy is to, every \(\Phi\) iterations, select for each layer \(l\) the top \(K\)\% of weights in \(W^l\) with the largest absolute changes since the last transport.
This can be formalized as:
\begin{equation}
    \begin{aligned}
        \Delta^l_{ij} & = \left|W^l_{ij} - \widetilde{W}^l_{ij} \right|, \\
        B^l_{ij}      & =
        \begin{cases}
            W^l_{ij} & \text{if } \Delta^l_{ij} \geq \tau^l \\
            B^l_{ij} & \text{o.w.}
        \end{cases},
    \end{aligned}
\end{equation}
where \(\widetilde{W}^l_{ij}\) is the value of the forward weight between neuron \(n^{l-1}_i\) and \(n^l_j\) at the time of its last transport, and \(\tau^l\) is the threshold corresponding to the top \(K\)\% of \(\Delta^l_{ij}\) values.
Determining \(\tau^l\) requires layer-level computation (not local to the synapse), and its complexity scales linearly with the number of weights.
This method effectively corrects larger biases but incurs additional computational and memory overhead in the forward network, as it involves finding \(\tau^l\) and storing previous weight values.

\paragraph{Random Sampling}
A low-complexity and hardware-friendlier alternative is to randomly sample weights for transport.
Every \(\Phi\) iterations, each weight in \(W\) is transported with a fixed probability \(P\) (Bernoulli sampling).
This method assumes that randomly sampling weights with a sufficiently high \(P\) is enough to adequately correct the bias over time.
However, it does not guarantee immediate correction of larger biases.
Since the probability is applied at the synapse level, it incurs no additional non-local computation or significant memory overhead.

\paragraph{Change-Weighted Sampling}
To correct larger biases while avoiding non-local computation, we build upon the two previous strategies by sampling weights based on the magnitude of their changes.
Unlike the previous strategies, this method does not transport the weights at a fixed interval \(\Phi\) but instead allows each forward weight to determine when to transport its value (hence, \(\Phi\) is set to 1).
The probability \(P^l_{ij}\) of transporting a weight \(W^l_{ij}\) after an iteration is:
\begin{equation}
    \begin{aligned}
        P^l_{ij} & = 1 - \exp\left(-\frac{\left|W^l_{ij} - \widetilde{W}^l_{ij} \right|}{\beta}\right), \\
        B^l_{ij} & =
        \begin{cases}
            W^l_{ij} & \text{with probability } P^l_{ij} \\
            B^l_{ij} & \text{o.w.}
        \end{cases},
    \end{aligned}
\end{equation}
where \(\beta\) is a temperature parameter controlling the sharpness of the selection probability.
\(\beta\) is a substitute for the fixed interval \(\Phi\) since it indirectly controls transport frequency: the lower \(\beta\), the more frequently weights are realigned.
Removing the need for a fixed interval allows this method to adapt more effectively to significant weight changes, which would otherwise require waiting for \(\Phi\) iterations before being corrected.
In addition, it simplifies the tuning process, as it only requires a single hyperparameter (\(\beta\)) instead of two (\(\Phi\), and \(K\) or \(P\)).
A drawback of this method is the additional memory required in the forward network to store previous weight values.

\section{Results} \label{chap:fbp:sec:results}

\subsection{Experimental Setup} \label{chap:fbp:sec:results-setup}

We describe the experimental setup used to evaluate our approach.
Additional details are provided in Appendix~\ref{chap:fbp:sec:exp-details}.
We select three standard image recognition datasets of increasing complexity: Fashion-MNIST~\citep{xiaoFashionMNISTNovelImage2017}, CIFAR-10~\citep{krizhevskyLearningMultipleLayers2009}, and CIFAR-100~\citep{krizhevskyLearningMultipleLayers2009}.
To ensure generalization across distinct \gls{snn} models, we evaluate our methods using both temporally-coded VGGs introduced in \citep{weiTemporalCodedSpikingNeural2023} (VGG-7 and VGG-11), and rate-coded ResNets introduced in~\citep{HuAdvancingSpikingResnet2024} (ResNet-18 and ResNet-26), which support deeper architectures compared to temporal coding.
The \gls{bp} algorithm is event-driven \gls{bp}~\citep{weiTemporalCodedSpikingNeural2023} for VGGs and \gls{bptt}~\citep{wuSpatioTemporalBackpropagationTraining2018} for ResNets.
These \gls{snn} baselines are selected due to their strong performance and the availability of open-source implementations; further details can be found in the respective references.
For training, we employ the Adam optimizer~\citep{kingmaAdamMethodStochastic2015}, L2 regularization, gradient clipping, early stopping, and annealing of the learning rate after each epoch.
Input images are normalized to \([0,1]\).
For CIFAR-10 and CIFAR-100, we employ simple data augmentation following the approach in~\citep{weiTemporalCodedSpikingNeural2023} to mitigate overfitting.
\(10\)\% of the training set is randomly reserved for validation (random holdout split).
Results on test sets are averaged over \(8\) trials with different random initializations; we report the mean and one standard deviation.
For each coding scheme, both architectural and training hyperparameters were optimized for \gls{bp} training on the validation set of CIFAR-10 using a gridsearch algorithm\footnote{We did not perform extensive tuning as our focus is relative performance rather than absolute accuracy.}.
Since all evaluated methods are built on top of \gls{bp}, using a shared configuration ensures a controlled and fair comparison.
We refer to Appendix~\ref{chap:fbp:sec:appendix:tc-exps-details} and~\ref{chap:fbp:sec:appendix:rc-exps-details} for hyperparameter values.

\subsection{Comparison with Existing Methods} \label{chap:fbp:sec:results-comparison}

We compare the performance of \gls{fbp} against \gls{bp}, along with \gls{sfa} and \gls{ss}.
In \gls{bp}, weight signs and values are transported after each iteration.
For \gls{sfa} and \gls{ss}, only weight signs are transported after each iteration.
In \gls{fbp}, \(\Phi\) is set to enable the transport of signs and values at a frequency of \(0.1\) (\(\Phi=10\)).
Thus, for a given epoch, weight transport occurs \(10\)\% of the time compared to its alternatives.
We report the highest \(\Phi\) maintaining, on average, optimal accuracy on the validation set of CIFAR-10.
For this comparison, we consider \gls{fbp} without partial weight transport.
Also, we focus on CIFAR-10 and CIFAR-100 in this section; results on Fashion-MNIST are reported in Appendix~\ref{chap:fbp:sec:appendix:results-comparison}.

\subsubsection{Temporal Coding} \label{chap:fbp:sec:tc-results-comparison}

We present the results for temporally-coded VGGs in Table~\ref{chap:fbp:tab:tc-acc-comp}.
Our method, \gls{fbp}, achieves a test accuracy comparable to \gls{bp} across all datasets and network architectures, while reducing weight transport per epoch by \(10\times\).
Note that \gls{bp} typically converges in fewer epochs on CIFAR-10 and CIFAR-100 (on average, \(9\)\% fewer epochs), leading to a minor reduction in total weight transports.
However, this reduction is largely offset by the substantial savings achieved per epoch in \gls{fbp}.
For instance, on CIFAR-10 with the VGG-7 architecture---where \gls{bp} requires \(11\)\% fewer epochs to converge---we measure \(46{,}112\) weight transports with \gls{bp} over \(262\) epochs, compared to only \(5{,}192\) with \gls{fbp} over \(295\) epochs (\(8.9\times\) reduction over the full training).

\begin{table}[t]
    \centering
    \setlength{\extrarowheight}{2pt}
    \footnotesize
    \caption{Accuracy comparison between our method, \gls{fbp}, and other feedback-driven methods, for training temporally-coded \glspl{snn}.}
    \begin{tabularx}{\linewidth}{@{}Ccccccc@{}}
        \toprule
        \multirow{2}{*}{Dataset}
                                       & \multirow{2}{*}{Architecture}
                                       & \multirow{2}{*}{Method}
                                       & \multicolumn{2}{c}{Transport Freq.}
                                       & \multirow{2}{*}{\makecell{Epochs                                                                                                                      \\(Mean\(\pm\)Std)}}
                                       & \multirow{2}{*}{\makecell{Accuracy                                                                                                                    \\(Mean\(\pm\)Std \%)}} \\

                                       &                                     &                                     & \makecell{Sign} & \makecell{Weight} &                &                    \\
        \midrule
        \multirow{8}{*}{CIFAR-10}      & \multirow{4}{*}{VGG-7}              & \glsentryshort{bp}                  & 1               & 1                 & \(262 \pm 34\) & \(90.40 \pm 0.45\) \\
                                       &                                     & \glsentryshort{sfa}                 & 1               & 0                 & \(298 \pm 50\) & \(87.03 \pm 0.69\) \\
                                       &                                     & \glsentryshort{ss}                  & 1               & 0                 & \(334 \pm 62\) & \(89.39 \pm 0.66\) \\
                                       &                                     & \glsentryshort{fbp} \textit{(ours)} & 0.1             & 0.1               & \(295 \pm 51\) & \(90.51 \pm 0.64\) \\
        \cmidrule(lr){2-7}
                                       & \multirow{4}{*}{VGG-11}             & \glsentryshort{bp}                  & 1               & 1                 & \(245 \pm 28\) & \(92.04 \pm 0.51\) \\
                                       &                                     & \glsentryshort{sfa}                 & 1               & 0                 & \(336 \pm 77\) & \(89.48 \pm 1.05\) \\
                                       &                                     & \glsentryshort{ss}                  & 1               & 0                 & \(291 \pm 55\) & \(91.26 \pm 0.53\) \\
                                       &                                     & \glsentryshort{fbp} \textit{(ours)} & 0.1             & 0.1               & \(270 \pm 59\) & \(92.04 \pm 0.49\) \\
        \midrule
        \multirow{8}{*}{CIFAR-100}     & \multirow{4}{*}{VGG-7}              & \glsentryshort{bp}                  & 1               & 1                 & \(308 \pm 29\) & \(66.05 \pm 0.42\) \\
                                       &                                     & \glsentryshort{sfa}                 & 1               & 0                 & \(377 \pm 61\) & \(60.97 \pm 0.92\) \\
                                       &                                     & \glsentryshort{ss}                  & 1               & 0                 & \(335 \pm 44\) & \(63.43 \pm 0.69\) \\
                                       &                                     & \glsentryshort{fbp} \textit{(ours)} & 0.1             & 0.1               & \(328 \pm 46\) & \(65.76 \pm 0.80\) \\
        \cmidrule(lr){2-7}
                                       & \multirow{4}{*}{VGG-11}             & \glsentryshort{bp}                  & 1               & 1                 & \(262 \pm 49\) & \(67.45 \pm 0.93\) \\
                                       &                                     & \glsentryshort{sfa}                 & 1               & 0                 & \(381 \pm 44\) & \(62.33 \pm 0.75\) \\
                                       &                                     & \glsentryshort{ss}                  & 1               & 0                 & \(289 \pm 41\) & \(65.34 \pm 0.79\) \\
                                       &                                     & \glsentryshort{fbp} \textit{(ours)} & 0.1             & 0.1               & \(291 \pm 27\) & \(67.25 \pm 0.62\) \\
        \bottomrule
    \end{tabularx}
    \label{chap:fbp:tab:tc-acc-comp}
\end{table}

\gls{fbp} outperforms existing feedback-driven methods relaxing weight symmetry through sign transport.
\gls{sfa} performs significantly worse than all other methods, confirming that stochasticity prevents scalability in deeper networks.
This aligns with prior studies on traditional \glspl{ann}~\citep{bartunovAssessingScalabilityBiologicallyMotivated2018,moskovitzFeedbackAlignmentDeep2019,xiaoBiologicallyPlausibleLearningAlgorithms2019}, though, to our knowledge, it has never been demonstrated in \glspl{snn}.
\gls{ss} provides a more effective alternative to \gls{sfa}, achieving accuracy closer to the \gls{bp} baseline.
However, due to the binary nature of the feedback weights, a performance gap remains and further widens on more challenging tasks like CIFAR-100.
In Appendix~\ref{chap:fbp:sec:appendix:results:tc-st}, we further study the impact of sign transport on our freezing mechanism.
We show that, unlike \gls{sfa} and \gls{ss}, enforcing sign symmetry during training with \gls{fbp} does not improve accuracy.

\subsubsection{Rate Coding} \label{chap:fbp:sec:rc-results-comparison}

We present the results for rate-coded ResNets in Table~\ref{chap:fbp:tab:rc-acc-comp}.
Again, \gls{fbp} remains close to \gls{bp} across all rate-coded architectures and datasets, with accuracy slightly below or above \gls{bp} depending on when early stopping is triggered (maximum drop of \(0.65\) pp below \gls{bp}).
Still, \gls{fbp} reduces total weight transport by \(8.3\)--\(12.5\times\) compared to \gls{bp}.

\begin{table}[t]
    \centering
    \setlength{\extrarowheight}{2pt}
    \footnotesize
    \caption{Accuracy comparison between our method, \gls{fbp}, and other feedback-driven methods, for training rate-coded \glspl{snn}.}
    \begin{tabularx}{\linewidth}{@{}Ccccccc@{}}
        \toprule
        \multirow{2}{*}{Dataset}
                                       & \multirow{2}{*}{Architecture}
                                       & \multirow{2}{*}{Method}
                                       & \multicolumn{2}{c}{Transport Freq.}
                                       & \multirow{2}{*}{\makecell{Epochs \\(Mean\(\pm\)Std)}}
                                       & \multirow{2}{*}{\makecell{Accuracy \\(Mean\(\pm\)Std \%)}} \\
                                       &                                     &                                     & \makecell{Sign} & \makecell{Weight} &                &                    \\
        \midrule
        \multirow{8}{*}{CIFAR-10}      & \multirow{4}{*}{ResNet-18}          & \glsentryshort{bp}                  & 1               & 1                 & \(514 \pm 66\)     & \(87.14 \pm 0.74\) \\
                                       &                                     & \glsentryshort{sfa}                 & 1               & 0                 & \(247 \pm 38\)     & \(55.29 \pm 0.81\) \\
                                       &                                     & \glsentryshort{ss}                  & 1               & 0                 & \(64 \pm 18\)      & \(10.80 \pm 1.08\) \\
                                       &                                     & \glsentryshort{fbp} \textit{(ours)} & 0.1             & 0.1               & \(516 \pm 117\)    & \(86.49 \pm 1.64\) \\
        \cmidrule(lr){2-7}
                                       & \multirow{4}{*}{ResNet-26}          & \glsentryshort{bp}                  & 1               & 1                 & \(365 \pm 66\)     & \(91.82 \pm 0.74\) \\
                                       &                                     & \glsentryshort{sfa}                 & 1               & 0                 & \(173 \pm 43\)     & \(55.90 \pm 1.20\) \\
                                       &                                     & \glsentryshort{ss}                  & 1               & 0                 & \(36 \pm 0\)       & \(10.00 \pm 0.00\) \\
                                       &                                     & \glsentryshort{fbp} \textit{(ours)} & 0.1             & 0.1               & \(392 \pm 64\)     & \(92.23 \pm 0.55\) \\
        \midrule
        \multirow{8}{*}{CIFAR-100}     & \multirow{4}{*}{ResNet-18}          & \glsentryshort{bp}                  & 1               & 1                 & \(434 \pm 67\)     & \(60.57 \pm 1.20\) \\
                                       &                                     & \glsentryshort{sfa}                 & 1               & 0                 & \(349 \pm 107\)    & \(31.87 \pm 1.00\) \\
                                       &                                     & \glsentryshort{ss}                  & 1               & 0                 & \(52 \pm 12\)      & \(0.90 \pm 0.09\) \\
                                       &                                     & \glsentryshort{fbp} \textit{(ours)} & 0.1             & 0.1               & \(526 \pm 85\)     & \(61.31 \pm 1.10\) \\
        \cmidrule(lr){2-7}
                                       & \multirow{4}{*}{ResNet-26}          & \glsentryshort{bp}                  & 1               & 1                 & \(326 \pm 59\)     & \(69.37 \pm 0.65\) \\
                                       &                                     & \glsentryshort{sfa}                 & 1               & 0                 & \(148 \pm 73\)     & \(28.03 \pm 1.19\) \\
                                       &                                     & \glsentryshort{ss}                  & 1               & 0                 & \(36 \pm 0\)       & \(1.00 \pm 0.00\) \\
                                       &                                     & \glsentryshort{fbp} \textit{(ours)} & 0.1             & 0.1               & \(260 \pm 73\)     & \(68.83 \pm 1.43\) \\
        \bottomrule
    \end{tabularx}
    \label{chap:fbp:tab:rc-acc-comp}
\end{table}

In contrast, existing feedback-driven methods fail to scale to deeper rate-coded ResNets.
\gls{sfa} incurs a severe accuracy drop on both datasets, while \gls{ss} collapses to near-chance performance.
Their poor performance is also reflected in the lower number of training epochs, which results from rapid convergence.
For \gls{ss}, even after tuning the hyperparameters and manually extending the number of epochs, we did not obtain satisfactory performance; see Appendix~\ref{chap:fbp:sec:appendix:results:ss-tuning} for details.
These results show that, for rate-coded deep \glspl{snn} based on \gls{bptt}, sign transport alone can be  insufficient to provide useful feedback alignment.
This degradation may arise because, unlike event-driven \gls{bp} in temporally-coded networks, \gls{bptt} propagates biased feedback errors across both depth and time, which may cause misalignment to accumulate more rapidly.
Overall, \gls{fbp} is the only feedback-driven method that preserves near \gls{bp}-level accuracy while reducing weight transport.
These results confirm that \gls{fbp} is suited to the most common coding schemes. 
Given the consistent trends observed across temporal and rate coding, the remaining analyses focus on temporally-coded networks.

\subsection{Impact of Partial Weight Transport} \label{chap:fbp:sec:results:tc-partial-wt}

\begin{figure}[ht]
    \centering
    \includegraphics[width=0.95\linewidth]{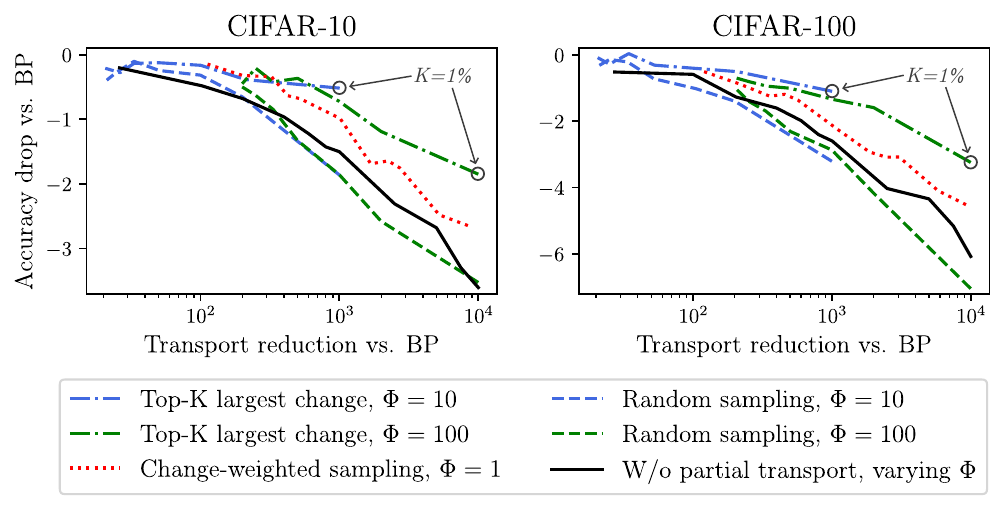}
    \caption{Accuracy drop versus weight transport reduction factor of \gls{fbp} relative to \gls{bp} on temporally-coded VGG-11, across different partial weight transport strategies and various numbers of frozen iterations \(\Phi\). Each strategy is evaluated by varying its specific hyperparameter. The x-axis uses a logarithmic scale, and y-axis ranges are adapted for each dataset to improve visualization. Best seen in color.}
    \label{chap:fbp:fig:tc-partial-wt}
\end{figure}

To reduce the total number of weights transported during training, two complementary methods can be employed: reducing the frequency of weight transports (via \(\Phi\), the number of iterations with frozen feedback weights), or reducing the number of weights transported at a time (via partial weight transport, controlled by a strategy-specific hyperparameter).
In Figure~\ref{chap:fbp:fig:tc-partial-wt}, we compare the accuracy drop against the weight transport reduction factor of \gls{fbp} relative to \gls{bp}, on temporally-coded VGG-11.
The analysis includes our three partial transport strategies and various values of \(\Phi\).
To ensure a fair comparison in terms of computational effort, we train each \gls{fbp} configuration for at most the number of epochs used by \gls{bp}, stopping training if convergence is not reached within this limit.
The transport reduction factor is computed as the ratio between the total number of weights transported in \gls{bp} and \gls{fbp}.
The total number of weights is obtained by multiplying the number of transports during training by the number of weights transferred per transport.
As a result, it depends on both \(\Phi\) and the specific hyperparameter of each partial weight transport strategy, which explains why the lines start and end at different positions along the x-axis.
Additional experimental details are provided in Appendix~\ref{chap:fbp:sec:appendix:tc-exps-details}, and supplementary results demonstrating the impact of the strategy-specific hyperparameters are reported in Appendix~\ref{chap:fbp:sec:appendix:results:tc-hyperparam}.

The solid black line shows the performance of the baseline: default \gls{fbp} without partial weight transport, across varying \(\Phi\) values.
For a given transport reduction factor, configurations that lie above this line achieve better accuracy at the same transport cost.
As observed, \textit{Random sampling} generally fails to outperform the \gls{fbp} baseline.
It shows marginal improvements only at low transport reduction levels, where either the probability of transporting each weight or the transport frequency is sufficiently high.
Its limited effectiveness can be explained by the stochastic nature of the sampling, which requires frequent realignment to reliably correct larger biases.
To address this, \textit{Change-weighted sampling} assigns higher transport probabilities to weights with larger accumulated changes, improving accuracy over \textit{Random sampling} and outperforming the \gls{fbp} baseline.
Also, it does not rely on a fixed transport interval (since \(\Phi=1\)), allowing weights to be realigned as needed.
The superior performance of this strategy highlights the importance of selection in partial weight transport.

Among all strategies, \textit{Top-K largest change} achieves the best performance.
This is because it explicitly corrects the largest weight biases.
It remains effective even at high transport reduction levels, with as few as \(K=1\)\% of weights transferred per transport.
Thus, on CIFAR-10, it can reduce weight transport by \(1{,}000\times\) with a \(0.52\)~pp accuracy drop (compared to \(1.50\)~pp for default \gls{fbp}), and by \(10{,}000\times\) with a \(1.85\)~pp drop (\(3.61\)~pp for default \gls{fbp}).
Similarly, on CIFAR-100, it achieves a \(1.10\)~pp drop with a \(1{,}000\times\) reduction (compared to \(2.60\)~pp for default \gls{fbp}), and a \(3.24\)~pp drop with a \(10{,}000\times\) reduction (\(6.08\)~pp for default \gls{fbp}).
Note that accuracy drops can be further reduced by training until early stopping: we measure a drop of \(1.15\)~pp on CIFAR-10 and \(1.06\)~pp on CIFAR-100 with an \(8{,}093\times\) and a \(7{,}003\times\) reduction, respectively.
While it remains unclear whether this strategy can be efficiently implemented on hardware, our goal is rather to demonstrate that transporting only a subset of the weights not only reduces the number of transports but also sufficiently corrects magnitude mismatches to support effective learning under relaxed weight symmetry.

\subsection{Impact of Weight Magnitude Mismatch} \label{chap:fbp:results:tc-wmismatch}
\begin{wrapfigure}{r}{0.425\linewidth}
    \centering%
    \includegraphics[width=1\linewidth]{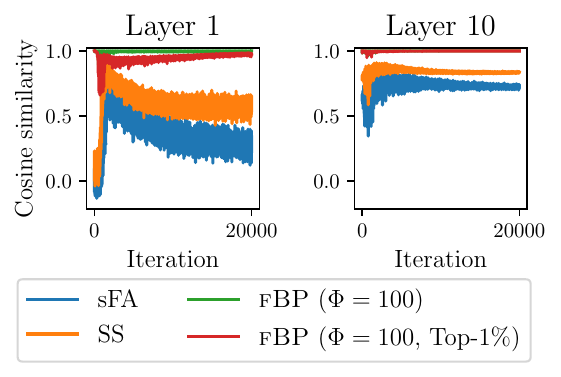}%
    \caption{Cosine similarity between true and actual weight changes during training on CIFAR-100 with VGG-11. }
    \label{chap:fbp:fig:cosim-dw}
\end{wrapfigure}
\gls{fbp} aims to minimize the gradient bias introduced by the magnitude mismatch between forward and feedback weights.
In this section, we quantify this bias by measuring the cosine similarity between true weight updates, computed using gradients based solely on forward weights (by replacing \(B\) with \(W\) in step~\ref{chap:fbp:alg:step-db} of Algorithm~\ref{chap:fbp:alg:bp-dualnetwork}), and actual weight updates, relying on neuron errors propagated through feedback weights.
Figure~\ref{chap:fbp:fig:cosim-dw} shows this alignment throughout training on CIFAR-100 with a temporally-coded VGG-11, across different feedback-driven methods.
\gls{fbp} maintains near-perfect alignment, even when combined with \textit{Top-K largest change} partial transport at \(K=1\)\%, consistently achieving higher cosine similarity compared to \gls{sfa} and \gls{ss}.
This alignment explains its superior performance, as gradients remain more accurate throughout training.
Interestingly, \gls{ss} shows higher similarity than \gls{sfa}, despite its feedback weights being limited to the sign of the forward weights.
This is also reflected empirically in its higher accuracy (for VGG-11), suggesting that the stochasticity in \gls{sfa} introduces noise in the gradient updates, leading to larger misalignment.
Last, cosine similarity tends to decrease in early layers.
This is because the magnitude mismatch between forward and feedback weights accumulates bias as gradients backpropagate through multiple layers.
\gls{fbp} is less affected than \gls{sfa} and \gls{ss} by this accumulated bias, illustrating its better scalability to deeper networks.
In Appendix~\ref{chap:fbp:sec:appendix:tc-wmismatch}, we provide additional results with similar conclusions across other datasets and all trainable layers.

\section{Discussion} \label{chap:fbp:sec:discussion}
In this paper, we improve training efficiency in dual-network configurations with our \gls{fbp} algorithm and our partial weight transport scheme.
Our results show that weight transport after every update is not necessary for near-optimal training, that transporting only a subset of the weights can sufficiently correct magnitude mismatches, and that enforcing sign symmetry during training provides no benefit when periodic weight transport is applied. These findings provide valuable insights to guide future efforts in the design of neuromorphic hardware with efficient \gls{bp}-based on-chip learning.


\gls{fbp} does not entirely eliminate the weight transport problem, as it still requires periodic synchronization.
Instead, it provides a middle ground between full weight symmetry and fully decoupled feedback weights, similar to \gls{sfa} and \gls{ss}, which rely on sign transport.
This can introduce design constraints, data movement overhead, and additional energy consumption on neuromorphic hardware.
While \gls{fbp} offers accuracy improvements over both \gls{sfa} and \gls{ss} and supports deeper \glspl{snn}, it requires additional transport of weight values along with the signs.
A drawback of \gls{sfa} and \gls{ss} is that they transmit signs after each iteration, which may incur higher synchronization overhead than \gls{fbp}.
The limited support for \gls{bp}-based deep learning in current neuromorphic hardware---often restricted to shallow architectures and local learning rules~\cite{rennerBackpropagationAlgorithmImplemented2024}---makes it difficult to reliably implement those methods and measure the energy cost of transporting weights or signs.
Thus, it remains uncertain whether more frequent sign transport offers a substantial efficiency advantage over weight transport.
At this stage, it is unclear which approach is best suited for \gls{bp}-based training on neuromorphic hardware.
The key challenge of \gls{fbp} is to find the optimal trade-off between transport frequency and accuracy that makes its overhead practical for hardware implementation.
This work does not claim our method to be the optimal solution, but rather seeks to provide insights for hardware–software co-design.

\gls{fbp} computes gradients using a stale version of the forward weights.
This approach raises similarities with methods employing delayed gradients~\citep{zhengAsynchronousStochasticGradient2017,zhuangFullyDecoupledNeural2022}, often encountered in asynchronous learning scenarios where workers compute gradients based on outdated model weights.
However, \gls{fbp} differs by separating forward and feedback weights (dual-network configuration), and by explicitly freezing the feedback weights for a given number of iterations (typically higher than in asynchronous learning).
Still, these characteristics make \gls{fbp} conceptually relevant to distributed~\citep{zhengAsynchronousStochasticGradient2017} and federated~\citep{zhuDelayedGradientAveraging2021} learning, which could inspire future work to draw from these domains, for example to mitigate the bias introduced by stale gradients~\citep{zhengAsynchronousStochasticGradient2017,zhuangFullyDecoupledNeural2022}.

\section*{Acknowledgements}
This work is funded by Chaire Luxant-ANVI (Métropole de Lille) and supported by IRCICA (CNRS UAR 3380).
Experiments presented in this paper were carried out using the Grid'5000 testbed~\citep{cappelloGrid5000LargeScale2005}, supported by a scientific interest group hosted by Inria and including CNRS, RENATER and several Universities as well as other organizations (see \url{https://www.grid5000.fr}).
We would like to thank Alpha Renner, Andrew Sornborger, and Gabriel Béna for the useful exchanges regarding the training of SNNs on neuromorphic hardware.
We also thank Wenjie Wei for providing the code from their paper, which our work builds upon.

\bibliographystyle{unsrtnat}
{\small
    \bibliography{references.bib}
}


\appendix

\section{Additional Results}
\renewcommand{\theequation}{A.\arabic{equation}}
\setcounter{equation}{0}
\renewcommand{\thefigure}{A.\arabic{figure}}
\setcounter{figure}{0}
\renewcommand{\thetable}{A.\arabic{table}}
\setcounter{table}{0}

\subsection{Experimental Details} \label{chap:fbp:sec:exp-details}

\subsubsection{Datasets} \label{chap:fbp:sec:exp-details-dataset}
We evaluate our methods on three datasets: Fashion-MNIST, CIFAR-10, and CIFAR-100.
Fashion-MNIST is a more challenging variant of MNIST~\citep{lecunGradientBasedLearningApplied1998}, comprising \(28\times28\) grayscale images, \(60{,}000\) samples for training and \(10{,}000\) for testing, categorized into \(10\) classes.
CIFAR-10 and CIFAR-100 contain \(32 \times 32\) RGB images, with \(50{,}000\) training samples and  \(10{,}000\) test samples.
They consist of \(10\) and \(100\) classes, respectively.
Fashion-MNIST is available at \url{https://github.com/zalandoresearch/fashion-mnist} under the MIT license.
CIFAR-10 and CIFAR-100 are available at \url{https://www.cs.toronto.edu/~kriz/cifar.html} under the MIT license.

\subsubsection{Setup for Temporal Coding} \label{chap:fbp:sec:appendix:tc-exps-details}
For temporally-coded networks, we employ the \gls{snn} model introduced in~\citep{weiTemporalCodedSpikingNeural2023}.
We evaluate our methods on two spiking VGG-based architectures: VGG-7 and VGG-11.
We do not consider deeper variants such as VGG-16 due to the increased difficulty of parameter initialization, which prior work typically addresses through \gls{ann} pretraining~\citep{weiTemporalCodedSpikingNeural2023,stanojevicHighPerformanceDeepSpiking2024}.
Our VGG-7 structure is \textit{64C3 -- 128C3 -- P2 -- 256C3 -- 256C3 -- P2 -- 512C3 -- 512C3 -- P2}, while VGG-11 is \textit{128C3 -- 128C3 -- 128C3 -- P2 -- 256C3 -- 256C3 -- 256C3 -- P2 -- 512C3 -- 512C3 -- 512C3 -- 512C3 -- P2}.
For convolutional layers (\textit{XC3}), \textit{X} denotes the number of channels and 3 the kernel size (\(3 \times 3\)). The stride is always \(1\), and zero-padding is applied to preserve spatial dimensions.
For pooling layers (\textit{P2}), we use max-pooling with a kernel of size \(2\), stride of \(2\), and no padding.
Both architectures are followed by a fully-connected layer, with the number of neurons matching the number of classes.
We use Kaiming initialization~\citep{heDelvingDeepRectifiers2015}, and do not add bias terms to reduce model complexity.
For training, we employ the Adam optimizer~\citep{kingmaAdamMethodStochastic2015} (\(\alpha = 10^{-4}\), \(\beta_1 = 0.9\), and \(\beta_2 = 0.999\)), L2 regularization (\(\lambda = 10^{-1}\)), gradient clipping (threshold of \(1\)), annealing of the learning rate (\(\alpha\)) after each epoch (factor of \(0.999\)), and early stopping with a patience of \(25\) epochs.
Unless specified otherwise, we use a batch size of \(256\).
Note that in the employed \gls{snn} model, time is continuous and no discrete time steps are used.
The maximum simulation duration is defined based on the number of layers (see \citep{weiTemporalCodedSpikingNeural2023}).

\paragraph{Details for Section~\ref{chap:fbp:sec:results:tc-partial-wt} and Appendix~\ref{chap:fbp:sec:appendix:results:tc-hyperparam}}
In these experiments, we reduce the batch size to \(32\) and adjust the initial learning rate to \(3\times10^{-5}\) (fine-tuned with gridsearch on the validation set of CIFAR-10, for \gls{bp} training).
Although this setting is more demanding in terms of computation time, it is motivated by two factors.
First, it provides a more realistic analysis by approximating an online learning scenario, which is more relevant for on-chip training on neuromorphic hardware.
Second, it allows us to validate our method under a different batch size, complementing the quantitative results in Section~\ref{chap:fbp:sec:results-comparison}, which used a batch size of \(256\).
We evaluate \textit{Change-weighted sampling} only with \(\Phi=1\) because the strategy operates without a fixed interval (see Section~\ref{chap:fbp:sec:methods-partial-wt}).
For \textit{Random sampling} and \textit{Top-K largest change}, we use \(\Phi=10\) and \(\Phi=100\), as these values provide strong baselines with minimal accuracy degradation in the absence of partial weight transport.
For their strategy-specific hyperparameters, we use the same ranges for \(P\) and \(K\) across all values of \(\Phi\), varying from \(0.5\) to \(0.01\) (with \(K\) expressed as a fraction); \(\beta_{\text{wt}}\) is evaluated over the range \(5\) to \(0.01\).
For \gls{fbp} without partial weight transport (baseline), we consider values of \(\Phi\) ranging from \(25\) to \(10{,}000\).

\subsubsection{Setup for Rate Coding} \label{chap:fbp:sec:appendix:rc-exps-details}
For rate-coded networks, we employ the \gls{snn} model introduced in~\citep{HuAdvancingSpikingResnet2024}.
We evaluate our methods on two spiking ResNet-based architectures: ResNet-18 and ResNet-26.
We do not consider deeper variants mainly due to the resources needed for thorough experiments and limited improvements observed on CIFAR-10. 
The general architecture is a sequence of residual layers (\textit{RL}), where each layer consists of multiple residual blocks. 
In all experiments, each residual layer contains two residual blocks.
Each residual block is composed of two convolution--batch normalization pairs, \textit{XC3 -- BN -- XC3 -- BN}, where \textit{X} denotes the number of output channels, \textit{C3} denotes a convolution with kernel size \(3 \times 3\), and \textit{BN} denotes batch normalization. 
A skip connection is included in the first residual block. 
Across consecutive residual layers, the number of channels is doubled.
We denote by \textit{XRL3} a residual layer with \(\textit{X}\) output channels, composed of two residual blocks with \(3 \times 3\) convolutions as described above. 
With this notation, the ResNet-18 architecture is given by \textit{16C3 -- 16RL3 -- 32RL3 -- 64RL3 -- 128RL3 -- FC}, and the ResNet-26 architecture is given by \textit{16C3 -- 16RL3 -- 32RL3 -- 64RL3 -- 128RL3 -- 256RL3 -- 512RL3 -- FC}.
The stride is always \(1\) and zero-padding is applied to preserve spatial dimensions except for the last one that uses a stride of two.
Both architectures are followed by a fully-connected layer, with the number of neurons matching the number of classes.
As recommended in the original paper~\citep{HuAdvancingSpikingResnet2024}, we use orthogonal initialization~\citep{xiao2018dynamical}, and do not add bias terms since batch normalization is present. 
For training, we employ the Adam optimizer~\citep{kingmaAdamMethodStochastic2015} (\(\alpha = 10^{-4}\), \(\beta_1 = 0.9\), and \(\beta_2 = 0.999\)), L2 regularization (\(\lambda = 10^{-3}\)), gradient clipping (threshold of \(1\)), annealing of the learning rate (\(\alpha\)) after each epoch (factor of \(0.999\)), early stopping with a patience of \(35\) epochs, and a batch size of \(256\).
The number of timesteps is set to 4.

\subsubsection{Computing Resources} \label{chap:fbp:sec:exp-details-resources}
Experiments were conducted on academic and private servers equipped with various Nvidia A100 GPUs (\(40\)~GiB) and running Debian Linux.
Code was implemented in Python~3.9 using PyTorch.
The total runtime depends on several factors, such as the dataset, architecture, batch size, and number of epochs.
For example, training a temporally-coded VGG-11 on CIFAR-100 with \gls{bp} took approximately \(15\) seconds per epoch with a batch size of \(256\) (\(6.9\)~GiB memory usage) and \(20\) seconds per epoch with a batch size of \(32\) (\(1.5\)~GiB memory usage).

\subsection{Impact of Sign Transport} \label{chap:fbp:sec:appendix:results:tc-st}

In this work, we employ periodic weight transport between the forward and the feedback weights to minimize magnitude mismatch and preserve gradient scale.
However, this approach can also be reduced to transport solely the signs of the weights.
In this section, we study the impact of sign transport on our freezing mechanism.
We perform this experiment with temporally-coded networks only, since \gls{ss} and \gls{sfa} fail with rate-coded networks.

\begin{figure}[ht]
    \centering
    \includegraphics[width=0.95\linewidth]{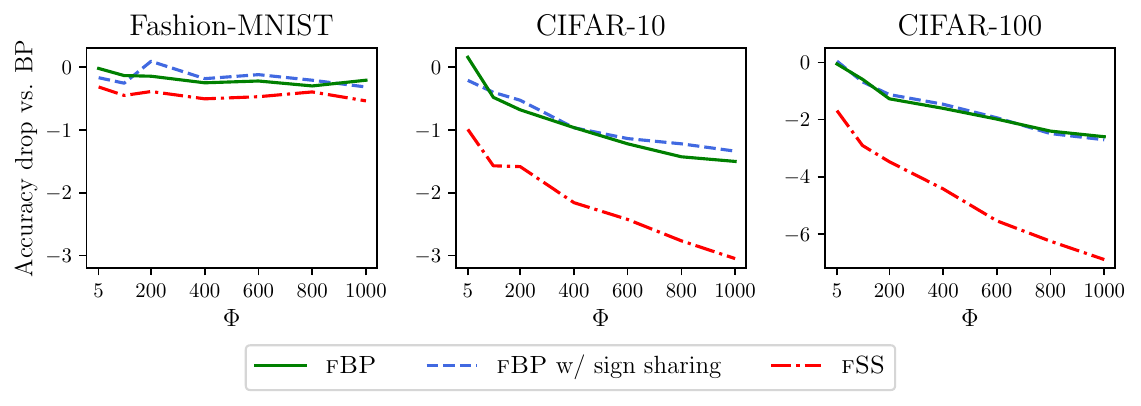}
    \caption{Accuracy drop of various methods with frozen feedback weights relative to \gls{bp}, evaluated on temporally-coded VGG-11 for varying number of iterations \(\Phi\). The y-axis range of the CIFAR-100 plot is larger to accommodate greater accuracy drops. Best seen in color.}
    \label{chap:fbp:fig:tc-impact-phi}
\end{figure}

In Figure~\ref{chap:fbp:fig:tc-impact-phi}, we evaluate, on the temporally-coded VGG-11 architecture, the accuracy drop of \gls{fbp}-based methods relative to \gls{bp}, across different values of \(\Phi\).
Each variant is introduced as it becomes relevant to the discussion.
To ensure a fair comparison in terms of computational effort, we train each \gls{fbp}-based method for the same number of epochs as \gls{bp}.
This implies that \gls{fbp} reduces weight transport by a factor of \(\Phi\) relative to \gls{bp}.
As in Section~\ref{chap:fbp:sec:results:tc-partial-wt}, we use a batch size of \(32\) and an initial learning rate of \(3\times10^{-5}\).

\paragraph{\textsc{fBP}} Globally, the need for frequent weight transport grows with the complexity of the task: on Fashion-MNIST, default \gls{fbp} (without partial weight transport) maintains near-optimal accuracy regardless of \(\Phi\), whereas on CIFAR-100, higher \(\Phi\) leads to larger accuracy drops.
Compared to Section~\ref{chap:fbp:sec:results-comparison}, \(\Phi\) can be increased with small performance degradation.
With \(\Phi=1000\), \gls{fbp} can reduce weight transport by \(1{,}000\times\) while maintaining accuracy within \(0.21\), \(1.50\), and \(2.60\)~pp of \gls{bp} on Fashion-MNIST, CIFAR-10, and CIFAR-100, respectively.
We demonstrated in Section~\ref{chap:fbp:sec:results:tc-partial-wt} that our partial weight transport scheme can further reduce accuracy drops under high transport reduction.

\paragraph{\textsc{fBP} with Sign Sharing} Prior work~\citep{liaoHowImportantWeight2016,xiaoBiologicallyPlausibleLearningAlgorithms2019} showed that enforcing sign symmetry is critical for scalability to deeper networks, motivating the development of \gls{sfa} and \gls{ss}.
To assess the specific contribution of sign symmetry in \gls{fbp}, we also evaluate it with sign sharing in Figure~\ref{chap:fbp:fig:tc-impact-phi}.
In this variant, the forward and feedback weights share the same signs, similarly to \gls{sfa} and \gls{ss}.
Specifically, at every iteration, the signs of the forward weights are copied to the feedback weights.
Weight transport (to correct magnitude mismatches) proceeds as in default \gls{fbp}, occurring at fixed intervals.
Note that sign sharing incurs additional transport costs not accounted for here, leading to a biased comparison but enabling a more accurate evaluation of its impact on performance.
Our results show that incorporating sign sharing has no significant effect on the accuracy achieved by \gls{fbp}, even at low transport frequencies (i.e., for higher values of \(\Phi\)).
This is likely because weight transport naturally corrects sign mismatches, although occurring at fixed intervals.
During these intervals, temporary sign mismatches may arise, but their impact remains limited.
Indeed, when a weight flips its sign during training, its magnitude is typically low both before and after the change.
Thus, forward and feedback weights may have opposite signs but remain small in magnitude, leading to negligible influence on both neuron activations and backpropagated errors.
We conclude that sign correction can be safely delayed until the next scheduled weight transport without affecting training.

\paragraph{\textsc{fSS}} We stated in the paper that freezing feedback weights is agnostic to the specific \gls{bp} algorithm.
Here, we evaluate the relevance of our freezing mechanism when applied to \gls{ss}.
We refer to this variant as \textsc{fSS} (frozen \gls{ss}).
In this case, weight signs are transported after \(\Phi\) iterations instead of every iteration, reducing the frequency of sign transport.
Figure~\ref{chap:fbp:fig:tc-impact-phi}  shows that \textsc{fSS} successfully supports training with our frozen feedback weights, illustrating again that our method can generalize beyond a specific \gls{bp}-based algorithm.
However, \gls{fbp} outperforms it across all datasets.
In addition, the accuracy of \textsc{fSS} degrades more rapidly than \gls{fbp} as \(\Phi\) increases.
This highlights that our freezing mechanism is less effective in \gls{ss} than in \gls{bp}, likely because \gls{ss} already imposes a strong constraint by limiting feedback weights to binary values, making it more sensitive to outdated feedback weights.
These results confirm that reducing magnitude mismatches between forward and feedback weights is important for mitigating gradient bias and achieving more effective learning in deep \glspl{snn}.

\subsection{Impact of Hyperparameters} \label{chap:fbp:sec:appendix:results:tc-hyperparam}

\begin{figure}[ht]
    \centering
    \includegraphics[width=0.9\linewidth]{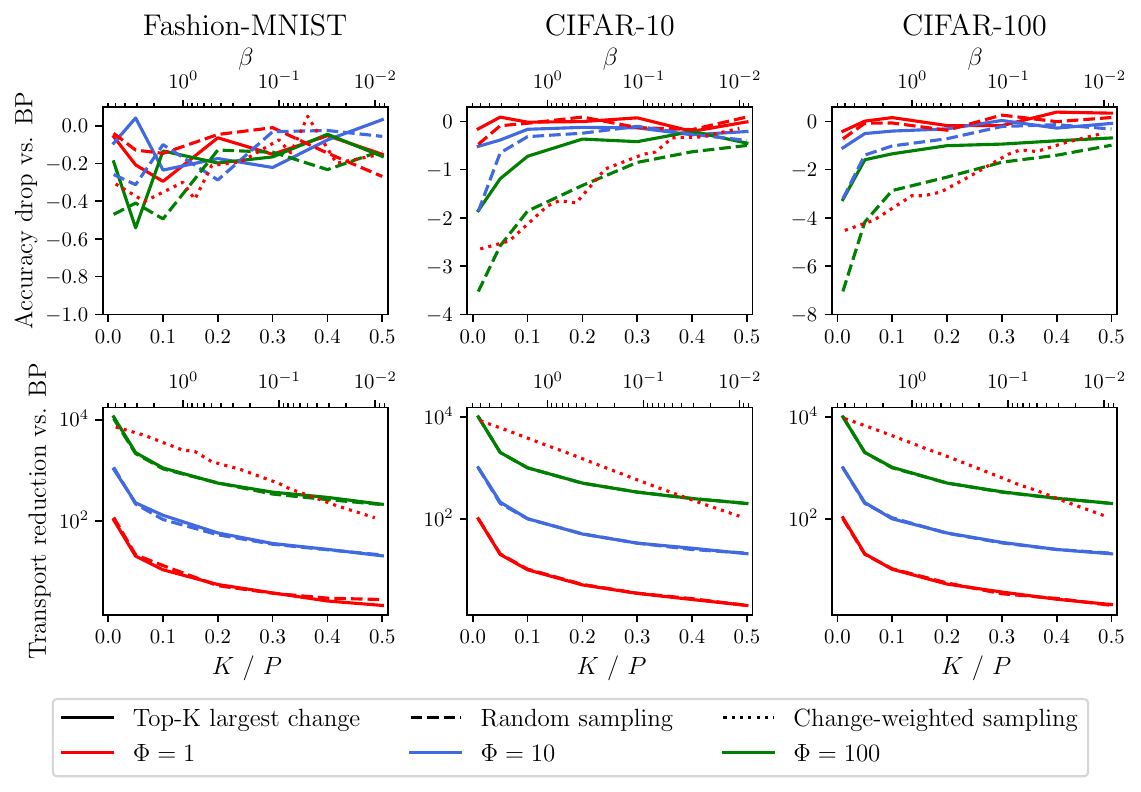}
    \caption{Accuracy drop versus weight transport reduction factor of \glsentryshort{fbp} relative to \glsentryshort{bp} on temporally-coded VGG-11, for different partial weight transport strategies and varying numbers of iterations \(\Phi\). The top x-axis is logarithmic, and the y-axis ranges differ for each dataset to enhance visualization. Best seen in color.}
    \label{chap:fbp:fig:tc-partial-wt-param}
\end{figure}

We showed in Section~\ref{chap:fbp:sec:results:tc-partial-wt} that the total number of weights transported during training can be reduced by lowering the transport frequency (via \(\Phi\)) and/or the number of weights transported at a time (via partial weight transport).
We introduced three partial weight transport strategies, each controlled by a distinct hyperparameter.
In this section, we provide additional visualizations across all datasets to illustrate the effect of these hyperparameters.
Figure~\ref{chap:fbp:fig:tc-partial-wt-param} presents the accuracy drop versus weight transport reduction factor of \gls{fbp} relative to \gls{bp}, for our different partial weight transport strategies.
\gls{fbp} is trained for the same number of epochs as \gls{bp}, ensuring a fair comparison.
This figure uses the same results as Figure~\ref{chap:fbp:fig:tc-partial-wt}, but splits them into separate plots for accuracy drop and transport reduction to better highlight the influence of the method-specific hyperparameters.
\textit{Top-K largest change} controls transport efficiency through \(\Phi\) and the top \(K\)\% of weights selected (bottom x-axis), \textit{Random sampling} through \(\Phi\) and a sampling probability \(P\) (bottom x-axis), and \textit{Change-weighted sampling} through a single temperature hyperparameter \(\beta\) (top x-axis).
For simplicity, we use the same ranges for \(P\) and \(K\) across all values of \(\Phi\).
We additionally include results for \(\Phi=1\) with \textit{Top-K largest change} and \textit{Random sampling}, compared to Figure~\ref{chap:fbp:fig:tc-partial-wt}.
\textit{Top-K largest change} and \textit{Random sampling} require joint tuning of their respective hyperparameters to effectively improve transport reduction, whereas \textit{Change-weighted sampling} relies on a single hyperparameter, making it easier to balance accuracy and efficiency.

\subsection{Comparison with Existing Methods} \label{chap:fbp:sec:appendix:results-comparison}

In Section~\ref{chap:fbp:sec:results-comparison}, we compared the performance of \gls{fbp} against \gls{bp}, along with \gls{sfa} and \gls{ss}, on CIFAR-10 and CIFAR-100. 
In this section, we provide the same results on the Fashion-MNIST dataset in Table~\ref{chap:fbp:tab:appendix:tc-acc-comp} for temporally-coded VGGs and in Table~\ref{chap:fbp:tab:appendix:rc-acc-comp} for rate-coded ResNets.
As observed, the results are consistent with the main findings: \gls{fbp} maintains similar performance to \gls{bp} while reducing weight transport by \(7.4\)--\(13.7\)\(\times\), depending on the total number of epochs relative to \gls{bp}.
It also outperforms \gls{sfa} and \gls{ss}, although the gap on VGGs is not significant because the simplicity of the task leads to accuracy saturation near the \gls{bp} level.
For rate-coded networks, \gls{sfa} degrades more substantially and \gls{ss} performs near chance, as observed on CIFAR-10 and CIFAR-100.

\begin{table}[ht]
    \centering
    \setlength{\extrarowheight}{2pt}
    \footnotesize
    \caption{Accuracy comparison between our method, \gls{fbp}, and other feedback-driven methods, for training temporally-coded \glspl{snn}.}
    \begin{tabularx}{\linewidth}{@{}Ccccccc@{}}
        \toprule
        \multirow{2}{*}{Dataset}
                                       & \multirow{2}{*}{Architecture}
                                       & \multirow{2}{*}{Method}
                                       & \multicolumn{2}{c}{Transport Freq.}
                                       & \multirow{2}{*}{\makecell{Epochs                                                                                                                      \\(Mean\(\pm\)Std)}}
                                       & \multirow{2}{*}{\makecell{Accuracy                                                                                                                    \\(Mean\(\pm\)Std \%)}} \\

                                       &                                     &                                     & \makecell{Sign} & \makecell{Weight} &                &                    \\
        \midrule
        \multirow{8}{*}{Fashion-MNIST} & \multirow{4}{*}{VGG-7}              & \glsentryshort{bp}                  & 1               & 1                 & \(110 \pm 37\) & \(92.98 \pm 0.21\) \\
                                       &                                     & \glsentryshort{sfa}                 & 1               & 0                 & \(91 \pm 23\)  & \(92.50 \pm 0.28\) \\
                                       &                                     & \glsentryshort{ss}                  & 1               & 0                 & \(102 \pm 33\) & \(92.75 \pm 0.22\) \\
                                       &                                     & \glsentryshort{fbp} \textit{(ours)} & 0.1             & 0.1               & \(104 \pm 32\) & \(92.99 \pm 0.27\) \\
        \cmidrule(lr){2-7}
                                       & \multirow{4}{*}{VGG-11}             & \glsentryshort{bp}                  & 1               & 1                 & \(104 \pm 28\) & \(92.98 \pm 0.21\) \\
                                       &                                     & \glsentryshort{sfa}                 & 1               & 0                 & \(89 \pm 14\)  & \(91.63 \pm 0.46\) \\
                                       &                                     & \glsentryshort{ss}                  & 1               & 0                 & \(94 \pm 20\)  & \(92.45 \pm 0.22\) \\
                                       &                                     & \glsentryshort{fbp} \textit{(ours)} & 0.1             & 0.1               & \(76 \pm 12\)  & \(92.90 \pm 0.16\) \\
        \bottomrule
    \end{tabularx}
    \label{chap:fbp:tab:appendix:tc-acc-comp}
\end{table}

\begin{table}[ht]
    \centering
    \setlength{\extrarowheight}{2pt}
    \footnotesize
    \caption{Accuracy comparison between our method, \gls{fbp}, and other feedback-driven methods, for training rate-coded \glspl{snn}.}
    \begin{tabularx}{\linewidth}{@{}Ccccccc@{}}
        \toprule
        \multirow{2}{*}{Dataset}
                                       & \multirow{2}{*}{Architecture}
                                       & \multirow{2}{*}{Method}
                                       & \multicolumn{2}{c}{Transport Freq.}
                                       & \multirow{2}{*}{\makecell{Epochs                                                                                                                      \\(Mean\(\pm\)Std)}}
                                       & \multirow{2}{*}{\makecell{Accuracy                                                                                                                    \\(Mean\(\pm\)Std \%)}} \\

                                       &                                     &                                     & \makecell{Sign} & \makecell{Weight} &                &                    \\
        \midrule
        \multirow{8}{*}{Fashion-MNIST} & \multirow{4}{*}{ResNet-18}          & \glsentryshort{bp}                  & 1               & 1                 & \(106 \pm 30\)     & \(91.46 \pm 0.33\) \\
                                       &                                     & \glsentryshort{sfa}                 & 1               & 0                 & \(127 \pm 41\)     & \(86.55 \pm 0.34\) \\
                                       &                                     & \glsentryshort{ss}                  & 1               & 0                 & \(56 \pm 15\)      & \(9.37 \pm 1.73\) \\
                                       &                                     & \glsentryshort{fbp} \textit{(ours)} & 0.1             & 0.1               & \(144 \pm 55\)     & \(91.58 \pm 0.39\) \\
        \cmidrule(lr){2-7}
                                       & \multirow{4}{*}{ResNet-26}          & \glsentryshort{bp}                  & 1               & 1                 & \(152 \pm 57\)     & \(92.55 \pm 0.28\) \\
                                       &                                     & \glsentryshort{sfa}                 & 1               & 0                 & \(152 \pm 48\)     & \(88.08 \pm 0.39\) \\
                                       &                                     & \glsentryshort{ss}                  & 1               & 0                 & \(36 \pm 0\)       & \(10.00 \pm 0.00\) \\
                                       &                                     & \glsentryshort{fbp} \textit{(ours)} & 0.1             & 0.1               & \(163 \pm 40\)     & \(92.77 \pm 0.29\) \\
        \bottomrule
    \end{tabularx}
    \label{chap:fbp:tab:appendix:rc-acc-comp}
\end{table}

\subsection{Impact of Weight Magnitude Mismatch} \label{chap:fbp:sec:appendix:tc-wmismatch}

\begin{figure}[ht]
    \centering
    \begin{subfigure}[b]{1\linewidth}
        \centering
        \includegraphics[width=1\linewidth]{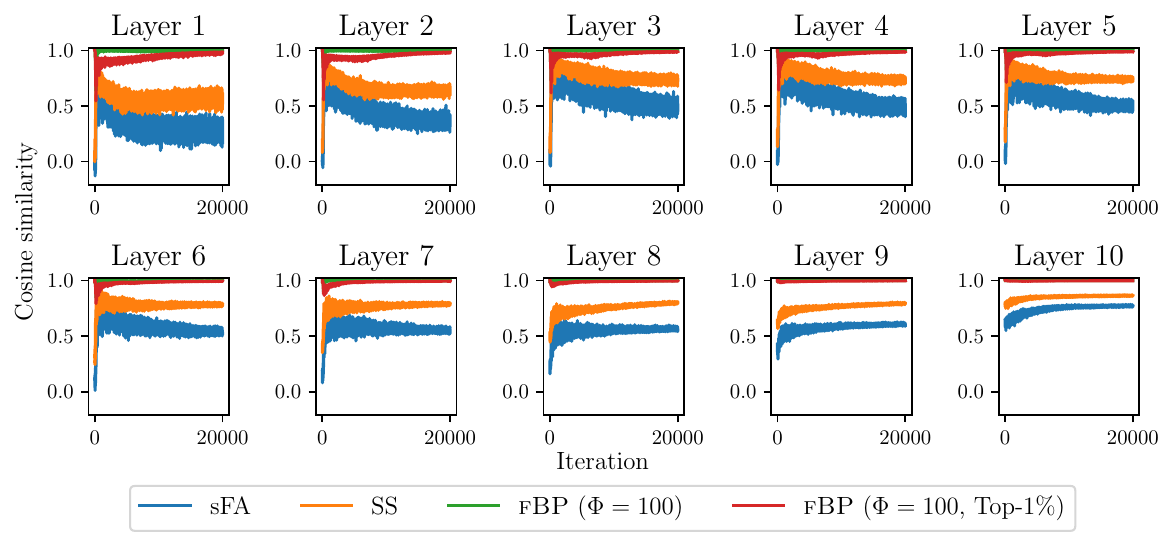}
        \caption{CIFAR-10}
        \label{chap:fbp:fig:cosim-dw-all-cifar10}
    \end{subfigure}
    \begin{subfigure}[b]{1\linewidth}
        \centering
        \includegraphics[width=1\linewidth]{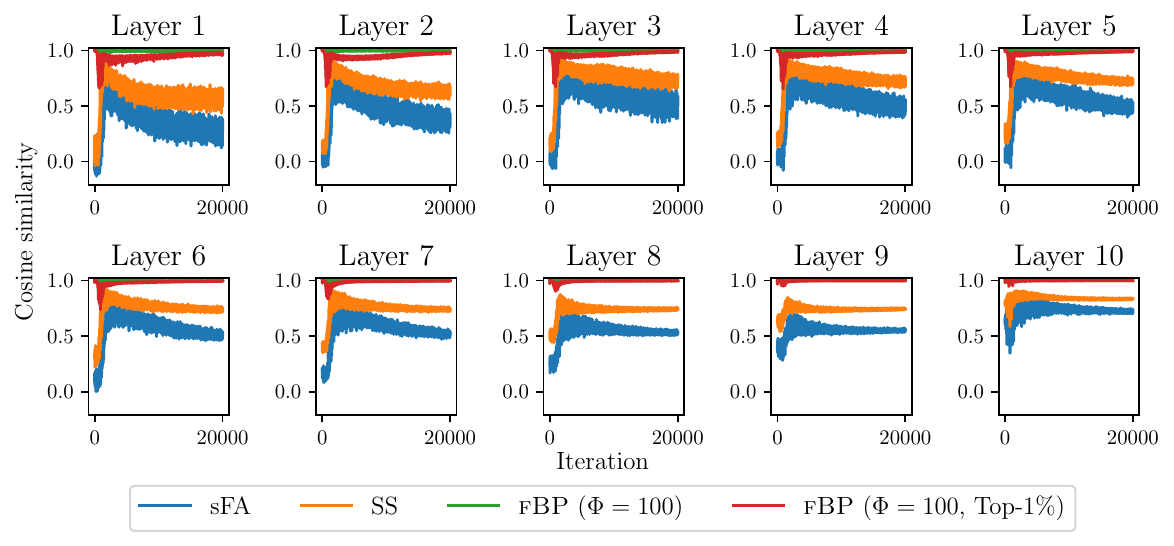}
        \caption{CIFAR-100}
        \label{chap:fbp:fig:cosim-dw-all-cifar100}
    \end{subfigure}
    \caption{Cosine similarity between true and actual weight changes during training with temporally-coded VGG-11. Best seen in color.}
    \label{chap:fbp:fig:cosim-dw-all}
\end{figure}

In Section~\ref{chap:fbp:results:tc-wmismatch}, we quantified the gradient bias introduced by the magnitude mismatch between forward and feedback weights.
To do so, we measured the cosine similarity between true weight updates, computed using gradients based solely on forward weights (i.e., by replacing \(B\) with \(W\) in step~\ref{chap:fbp:alg:step-db} of Algorithm~\ref{chap:fbp:alg:bp-dualnetwork}), and actual weight updates, relying on neuron errors propagated through feedback weights.
We showed that \gls{fbp} maintains near-perfect alignment, consistently achieving higher cosine similarity compared to \gls{sfa} and \gls{ss}.
Here, we provide additional figures on CIFAR-10 (Figure~\ref{chap:fbp:fig:cosim-dw-all-cifar10}) and CIFAR-100 (Figure~\ref{chap:fbp:fig:cosim-dw-all-cifar100}) across all the trainable layers of a temporally-coded VGG-11.
Note that the last fully-connected layer is excluded because all methods receive the same neuron errors, resulting in identical weight changes.
These results highlight that cosine similarity tends to decrease in early layers due to accumulated gradient bias during the \gls{bp} phase.
\gls{fbp} is less impacted than \gls{sfa} and \gls{ss} by this accumulated bias, which explains its better scalability to deeper networks.

\subsection{Extended Tuning of Feedback-Driven Baselines} \label{chap:fbp:sec:appendix:results:ss-tuning}
In Section~\ref{chap:fbp:sec:rc-results-comparison}, we compared our \gls{fbp} algorithm with \gls{ss} on rate-coded ResNets. 
Using the hyperparameters tuned for \gls{bp}, \gls{ss} failed to converge on CIFAR-10 and CIFAR-100. 
To assess whether this failure was caused by the method rather than the hyperparameter choice, we conducted an extended hyperparameter search over 250 configurations on CIFAR-10 and CIFAR-100.
We varied the batch size (\(32\), \(256\)), learning rate (\(3 \times 10^{-5}\) to \(1 \times 10^{-2}\)), weight decay (\(1 \times 10^{-4}\) to \(1 \times 10^{-1}\)), early-stopping patience, and ResNet depth. 
We increased the patience up to 200 epochs to ensure that early stopping was not triggered before the models had sufficient time to train.
Despite this tuning, \gls{ss} remained far below \gls{bp} and \gls{fbp}, with validation accuracy ranging from \(9.32\)\% to \(67.82\)\% on CIFAR-10, and from \(0.88\)\% to \(15.97\)\% on CIFAR-100.
These results confirm that \gls{ss} does not extend well to deeper rate-coded architectures.


\section{Neuron Errors in BPTT-Based Training} \label{chap:fbp:sec:bptt-neuron-errors}
\renewcommand{\theequation}{B.\arabic{equation}}
\setcounter{equation}{0}
\renewcommand{\thefigure}{B.\arabic{figure}}
\setcounter{figure}{0}
\renewcommand{\thetable}{B.\arabic{table}}
\setcounter{table}{0}

Neuron errors are the central concept for characterizing \gls{bp} training in dual-network configurations.
In the main text, we defined neuron errors for event-driven \gls{bp}. 
Here, we provide the corresponding formulation for \gls{bptt}.

In \gls{bptt}, the \gls{snn} dynamics are unfolded over discrete time steps. 
Let \(s^l\left[t\right]\) denote the spike output of layer \(l\) at time step \(t\), and let \(u^l\left[t\right]\) denote the corresponding membrane potential. 
The loss gradient with respect to the weights of layer \(l\) is obtained by accumulating contributions over all time steps:
\begin{equation}
    \frac{\partial \mathcal{L}}{\partial W^l}
    =
    \sum_{t=1}^{T}
    \underbrace{
        \frac{\partial \mathcal{L}}{\partial s^l\left[t\right]}
    }_{\delta^l\left[t\right]}
    \frac{\partial s^l\left[t\right]}{\partial u^l\left[t\right]}
    \frac{\partial u^l\left[t\right]}{\partial W^l}.
    \label{chap:fbp:eq:grad-dldw-bptt}
\end{equation}
where \(T\) is the number of simulation time steps, \(W^l\) are the weights of layer \(l\), and \(\mathcal{L}\) is the loss.
The error term \(\delta^l\left[t\right] = \frac{\partial \mathcal{L}}{\partial s^l\left[t\right]}\)
captures the gradient of the loss with respect to the spike output of neurons in layer \(l\) at time \(t\).
This is what we refer to as the neuron errors in \gls{bptt}.
As in event-driven \gls{bp}, these errors represent the non-local feedback signal propagated from deeper layers.
The remaining factors,
\(\frac{\partial s^l\left[t\right]}{\partial u^l\left[t\right]}\) and
\(\frac{\partial u^l\left[t\right]}{\partial W^l}\),
capture the local contribution to the gradient.

\end{document}